\documentclass[10pt,twocolumn,letterpaper]{article}

\usepackage{cvpr}              %

\usepackage{graphicx}
\usepackage{amsmath}
\usepackage{amssymb}
\usepackage{booktabs}
\usepackage[]{algorithm2e}
\usepackage{times}
\usepackage{epsfig}
\usepackage{amsfonts}
\usepackage{amsthm}
\usepackage{url}
\usepackage{color}
\usepackage{xcolor}
\usepackage{bbm}
\usepackage{tabularx}
\usepackage{xspace}
\usepackage{paralist} %
\usepackage{caption}
\usepackage{booktabs}
\usepackage{microtype}
\usepackage{wrapfig}
\usepackage{adjustbox}
\usepackage{multirow}
\usepackage{afterpage}
\usepackage{makecell}
\usepackage{cuted}
\usepackage{capt-of}

\usepackage[font=small]{caption}
\usepackage[outline]{contour}

\newcommand{\Ia}{\mathbf{I}_{0}}
\newcommand{\Ib}{\mathbf{I}_{1}}
\newcommand{\intT}{t}

\newcommand{\LDIa}{\mathcal{L}_{0}}
\newcommand{\LDIb}{\mathcal{L}_{1}}

\newcommand{\FLDIa}{\mathcal{F}_{0}}
\newcommand{\FLDIb}{\mathcal{F}_{1}}

\newcommand{\sfa}{\mathbf{u}_{0}}
\newcommand{\sfb}{\mathbf{u}_{1}}

\newcommand{\ColorLa}{\mathbf{C}_{0}^l}
\newcommand{\ColorLb}{\mathbf{C}_{1}^l}

\newcommand{\DepthLa}{\mathbf{D}_{0}^l}
\newcommand{\DepthLb}{\mathbf{D}_{1}^l}

\newcommand{\FeatureLa}{\mathbf{F}_{0}^l}
\newcommand{\FeatureLb}{\mathbf{F}_{1}^l}

\newcommand{\Featurea}{\mathbf{f}_{0}}
\newcommand{\Featureb}{\mathbf{f}_{1}}

\newcommand{\Posa}{\mathbf{x}_{0}}

\newcommand{\Posat}{\mathbf{x}_{0 \rightarrow t}}
\newcommand{\Posbt}{\mathbf{x}_{1 \rightarrow t}}

\newcommand{\ignore}[2]{\hspace{0in}#2}

\newcommand{\Posb}{\mathbf{x}_{1}}

\newcommand{\Depthat}{\mathbf{D}_{0 \rightarrow t}}
\newcommand{\Depthbt}{\mathbf{D}_{1 \rightarrow t}}
\newcommand{\Deptht}{\mathbf{D}_{t}}

\newcommand{\FeatureMapat}{\mathbf{F}_{0 \rightarrow t}}
\newcommand{\FeatureMapbt}{\mathbf{F}_{1 \rightarrow t}}
\newcommand{\FeatureMapt}{\mathbf{F}_{t}}

\newcommand{\WeightMapt}{\mathbf{W}_{t}}

\newcommand{\Ptsa}{\mathcal{P}_{0}}
\newcommand{\Ptsb}{\mathcal{P}_{1}}

\usepackage[pagebackref,breaklinks,colorlinks]{hyperref}

\usepackage[capitalize]{cleveref}
\crefname{section}{Sec.}{Secs.}
\Crefname{section}{Section}{Sections}
\Crefname{table}{Table}{Tables}
\crefname{table}{Tab.}{Tabs.}

\def\cvprPaperID{7814} %
\def\confName{CVPR}
\def\confYear{2022}

\begin{document}

\title{3D Moments from Near-Duplicate Photos}
\author{
Qianqian Wang$^{1, 2}$ \ \  
Zhengqi Li$^1$ \ \
David Salesin$^1$ \ \
Noah Snavely$^{1,2}$ \ \
Brian Curless$^{1,3}$ \ \  
Janne Kontkanen$^1$  
\\[0.5em]
$^1$Google Research \ \ \
$^2$Cornell Tech, Cornell University \ \ \ $^3$University of Washington
\ \ \
}
\maketitle

\begin{strip}\centering
\includegraphics[width=\textwidth]{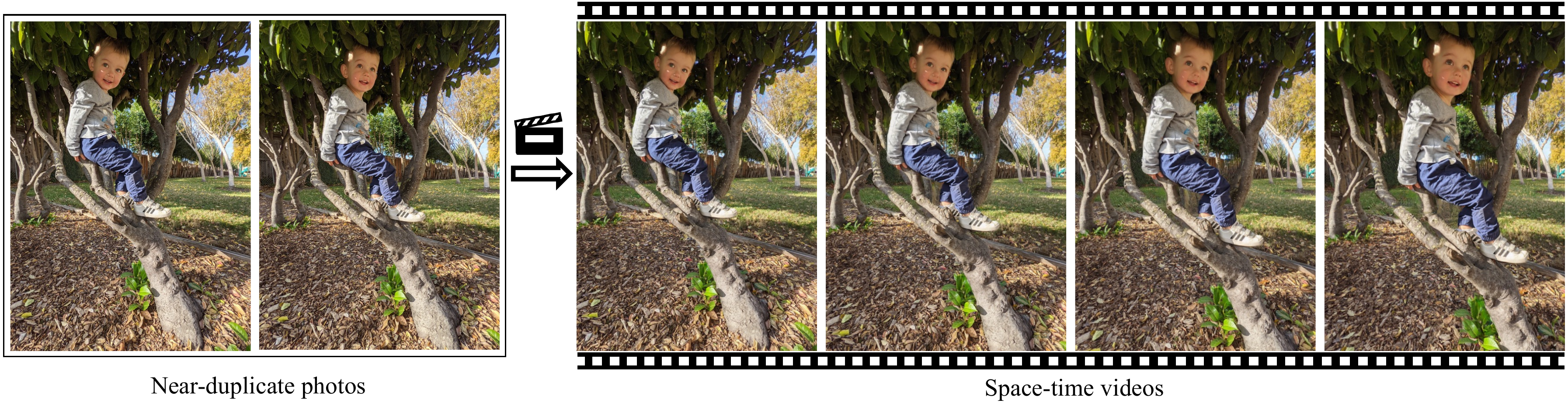} 
\vspace{-1em}
\captionof{figure}{
People often take many near-duplicate photos in an attempt to capture the perfect expression. Given a pair of these photos, taken from nearby viewpoints~(\textbf{left}), our proposed approach brings these photos to life as 3D Moments, producing space-time videos with cinematic camera motions and interpolated scene motion~(\textbf{right}). Please refer to the supplementary material to see the videos. 
\label{fig:teaser}} 
\end{strip}

\begin{abstract}
We introduce {\em 3D Moments}, a new computational photography effect. 
As input we take a pair of near-duplicate photos, i.e., photos of moving subjects from similar viewpoints, common in people's photo collections. As output, we produce a video that smoothly interpolates the scene motion from the first photo to the second, while also producing camera motion with parallax that gives a heightened sense of 3D.
To achieve this effect, we represent the scene as a pair of feature-based layered depth images augmented with scene flow.  This representation enables motion interpolation 
along with independent control of the camera viewpoint. 
Our system produces photorealistic space-time videos with motion parallax and scene dynamics, while plausibly recovering regions occluded in the original views. We conduct extensive experiments demonstrating superior performance over baselines on public datasets and in-the-wild photos. Project page: \url{https://3d-moments.github.io/}.

\end{abstract}
\section{Introduction}

Digital photography enables us to take scores of photos in order to capture just the right moment. 
In fact, we often end up with many near-duplicate photos in our image collections as we try to capture the best facial expression of a family member, or the most memorable part of an action. 
These near-duplicate photos end up just lying around in digital storage, unviewed.

In this paper, we aim to utilize such near-duplicate photos to create a compelling new kind of 3D photo enlivened with animation. 
We call this new effect \emph{3D Moments}: given a pair of near-duplicate photos depicting a dynamic scene 
from nearby (perhaps indistinguishable) viewpoints, such as the images in Fig.~\ref{fig:teaser} (left), our goal is to simultaneously enable cinematic camera motion with 3D parallax (including novel, extrapolated viewpoints) while faithfully interpolating scene motion
to synthesize short space-time videos like the one shown in Fig.~\ref{fig:teaser} (right). 
3D Moments combine both camera and scene motion in a compelling way,
but involve very challenging vision problems: we must jointly infer 3D geometry, scene dynamics, and content that becomes newly disoccluded during the animation.

Despite great progress towards each of these individual problems, tackling all of them jointly is non-trivial, especially with image pairs with unknown camera poses as input. NeRF-based view synthesis methods for dynamic scenes~\cite{Park2020DeformableNR, Xian2021SpacetimeNI, Li2021NeuralSF, Park2021HyperNeRFAH} require many images with known camera poses.
Single-photo view synthesis methods (sometimes called 3D Photos or 3D Ken Burns~\cite{Niklaus20193DKB, shih20203d, Kopf2020OneS3}) can create animated camera paths from a single photo, but cannot represent moving people or objects. Frame interpolation can create smooth animations from image pairs, but only in 2D. Furthermore, naively applying view synthesis and frame interpolation methods sequentially results in temporally inconsistent, unrealistic animations.

To address these challenges, we propose a novel approach for creating 3D Moments by explicitly modeling time-varying geometry and appearance from two uncalibrated, near-duplicate photos. The key to our approach is to represent the scene as a pair of feature-based layered depth images (LDIs) augmented with scene flows. We build this representation by first transforming the input photos into a pair of color LDIs, with inpainted color and depth for occluded regions.  We then extract features for each layer with a neural network to create the feature LDIs.  In addition, we compute optical flow between the input images and combine it with the depth layers to estimate scene flow between the LDIs. To render a novel view at a novel time, we lift these feature LDIs into a pair of 3D point clouds, and employ a depth-aware, bidirectional splatting and rendering module that combines the splatted features from both directions.

We extensively test our method on both public multi-view dynamic scene datasets and in-the-wild photos in terms of rendering quality, and demonstrate superior performance compared to state-of-the-art baselines. 

In summary, our main contributions include: 
(1) the new task of creating 3D Moments from near-duplicate photos of dynamic scenes, and
(2) a new representation based on feature LDIs augmented with scene flows, and a model that can be trained for creating 3D Moments.

\section{Related work}

Our work builds on methods for 
few-shot view synthesis, 
frame interpolation and space-time view synthesis. 

\smallskip
\noindent\textbf{View synthesis from one or two views.} 
Novel view synthesis aims to reconstruct unseen viewpoints from a set of input 2D images. Recent neural rendering methods achieve impressive synthesis results~\cite{mildenhall2020nerf,wang2021ibrnet,liu2020neural,tewari2020NeuralSTAR,tewari2021advances,zhang2020nerf++}, but typically assume many views as input and thus do not suit our task. We focus here on methods that take just one or two views.
Many single-view synthesis methods involve estimating dense monocular depths and filling in occluded regions~\cite{li2019learning,Niklaus20193DKB,Wiles2020SynSinEV,shih20203d,Kopf2020OneS3,Rockwell2021,jampani2021SLIDESI}, while others seek to directly regress to a scene representation in a single step~\cite{single_view_mpi,tulsiani2018layer,rafique2020gaf,rombach2021geometry,yu2021pixelnerf}. We draw on ideas from several works in this vein: SynSin learns a feature 3D point cloud for each input image and projects it to the target view where the missing regions are inpainted~\cite{Wiles2020SynSinEV}. 3D Photo~\cite{shih20203d} instead creates a Layered Depth Image~(LDI) and inpaints the color and depth of the occluded region in a spatial context-aware manner. We build on both methods but extend to the case of dynamic scenes. 

Like our method, some prior view synthesis methods operate on two views. For instance, Stereo Magnification~\cite{zhou2018stereo} and related work~\cite{srinivasan2019pushing} take two narrow-baseline stereo images and predict a multi-plane image that enables real-time novel view synthesis.
However, unlike our approach, these methods assume that there is some parallax from camera motion, and again can only model static scenes, not ones where there is scene motion between the two input views.

\smallskip
\noindent\textbf{Frame interpolation.}
In contrast to 3D view synthesis, temporal frame interpolation creates sequences of in-between frames from two input images.  Frame interpolation methods do not distinguish between camera and scene motion: all object motions are interpolated in 2D image space.  Moreover, most frame interpolators assume a linear motion model~\cite{Niklaus_CVPR_2017_adaconv, Bao_CVPR_2019_dain, Niklaus_ICCV_2017_sepconv, Niklaus_ARXIV_2020_resepconv, lee2020adacof, Niklaus_CVPR_2020_softsplat,  park2020bmbc, Park2021AsymmetricBM, Sim2021XVFIEV, Jiang_CVPR_2018_super} although some recent works consider quadratic motion~\cite{xu2019quadratic, liu2020enhanced}. Most of the interpolators use image warping with optical flow, 
although as a notable exception, Niklaus et al.~\cite{Niklaus_ICCV_2017_sepconv, Niklaus_ARXIV_2020_resepconv} synthesize intermediate frames by blending the inputs with kernels predicted by a neural network. However, frame interpolation alone cannot generate 3D Moments, since it does not recover the 3D geometry or allow control over camera motion in 3D.

\medskip
\noindent\textbf{Space-time view synthesis.}
A number of methods have sought to synthesize novel views for dynamic scenes in both space and time by modeling time-varying 3D geometry and appearance. 
Many methods require synchronized multi-view videos as inputs, and thus do not apply to in-the-wild photos~\cite{bansal20204d, Bemana2020xfields, stich2008view, zitnick2004high, Li2021Neural3V,broxton2020immersive}. Recently, several neural rendering approaches~\cite{yoon2020novel, Li2021NeuralSF, Xian2021SpacetimeNI, Park2020DeformableNR, Pumarola2021DNeRFNR, Park2021HyperNeRFAH} have shown promising results on space-time view synthesis from monocular dynamic videos. To interpolate both viewpoints and time, recent works either directly interpolate learned latent codes~\cite{Park2020DeformableNR, Park2021HyperNeRFAH}, or apply splatting with estimated 3D scene flow fields~\cite{Li2021NeuralSF}. However, these methods require densely sampled input views with accurate camera poses, which are unavailable for our two-image setting. Moreover, none of them explicitly inpaint unseen regions.

\section{Method}
\begin{figure*}[h]
    \centering
    \includegraphics[width=\linewidth]{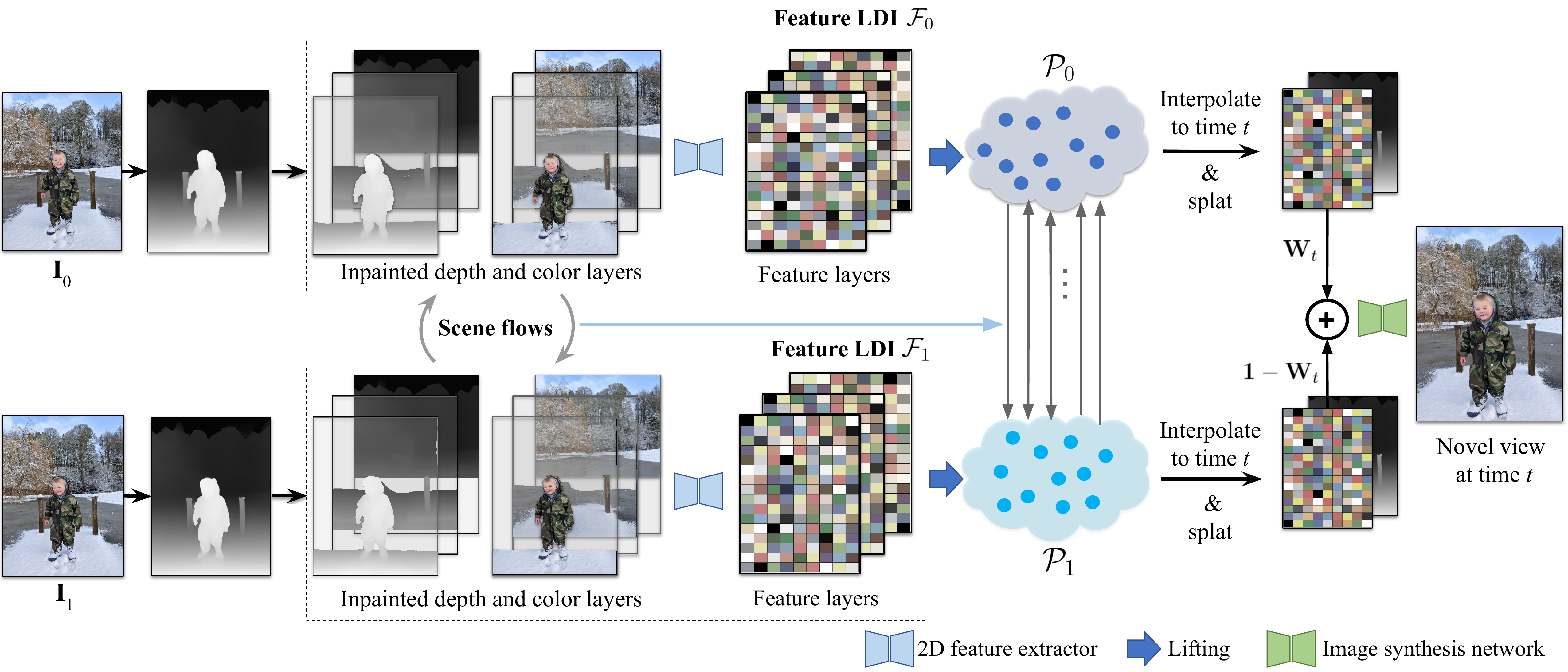} \vspace{-0.5em}
    \caption{\textbf{Overview.} Given near-duplicate photos ($\Ia$, $\Ib$), we align them with a homography and predict a dense depth map for each photo. Each RGBD image is then converted to a color LDI, with  occluded regions filled by depth-aware inpainting. A 2D feature extractor is applied to each color layer of the inpainted LDIs to obtain feature layers, resulting in feature LDIs ($\FLDIa, \FLDIb$), where colors in the inpainted LDIs have been replaced with features. 
    To model scene motion, we compute the scene flow of each pixel in the LDIs using the predicted depths and optical flows between the two input images. To render a novel view at intermediate time $t$, we lift the feature LDIs to a pair of 3D point clouds ($\Ptsa$, $\Ptsb$) and bidirectionally move points along their scene flows to time $t$.  We then project and splat these 3D feature points to form forward and backward 2D feature maps (from $\Ptsa$ and $\Ptsb$, respectively) and their corresponding depth maps. We linearly blend these maps with weight map $\mathbf{W}_t$ derived from spatio-temporal cues, and pass the result to an image synthesis network to produce the final image.}
    \label{fig:overview}
\end{figure*}

\subsection{Problem statement and method overview}
\label{sec3.1:problem_defination}
The input to our system is a pair of images $(\Ia, \Ib)$ of a dynamic scene taken at nearby times and camera viewpoints. For tractable motion interpolation, we assume that motion between $\Ia$ and $\Ib$ is roughly within the operating range of a modern optical flow estimator.
Our goal is to create 3D Moments by independently controlling the camera viewpoint while simultaneously interpolating scene motion to render arbitrary nearby novel views at arbitrary intermediate times $\intT \in [0, 1]$. Our output is a space-time video with cinematic camera motions and interpolated scene motion.

To this end, we propose a new framework that enables efficient and photorealistic space-time novel view synthesis without the need for test-time optimization. Our pipeline is illustrated in Fig.~\ref{fig:overview}. Our system starts by aligning the two photos into a single reference frame via a homography. The key to our approach is building feature LDI from each of the inputs, where each pixel in the feature LDI consists of its depth, scene flow and a learnable feature. 

To do so, we first transform each input image into a color LDI~\cite{shade1998layered} with inpainted color and depth in occluded regions.
We then extract deep feature maps from each color layer of these LDIs to obtain a pair of {\rm feature} LDIs ($\FLDIa, \FLDIb$). To model scene dynamics, the scene flows of each pixel in the LDIs are estimated based on predicted depth and optical flows between the two inputs. 
Finally, to render a novel view at intermediate time $t$, we lift the feature LDIs into a pair of point clouds $(\Ptsa, \Ptsb)$ and propose a scene-flow-based bidirectional splatting and rendering module to combine the features from two directions and synthesize the final image. We now describe our method in more detail.

\subsection{LDIs from near-duplicate photos}
\label{sec3.2:LDIs}

Our method first computes the underlying 3D scene geometry. As near-duplicates typically have scene dynamics and very little camera motion, standard Structure from Motion (SfM) and stereo reconstruction methods fail to produce reliable results. Instead, we found that state-of-the-art monocular depth estimator DPT~\cite{Ranftl2021} can produce sharp and plausible dense depth maps for images in the wild. Therefore, we rely on DPT to obtain the geometry for each image.

To account for small camera 
pose changes between the views, we compute optical flow between the views using RAFT~\cite{teed2020raft}, estimate a homography between the images using the flow, and then warp $\Ib$ to align with $\Ia$. Because we only want to align the static background of two images, we mask out regions with large optical flow, which often correspond to moving objects, and compute the homography using the remaining mutual correspondences given by the flow. 
Once $\Ib$ is warped to align with $\Ia$, %
we treat their camera poses as identical.
To simplify notation, we henceforth re-use $\Ia$ and $\Ib$ to denote the aligned input images.

We then apply 
DPT~\cite{Ranftl2020} 
to predict the depth maps for each image.
To align the depth range of $\Ib$ with $\Ia$ we estimate a global scale and shift for $\Ib$'s disparities (i.e., 1/depth), using flow correspondences in the static regions. 
Next, we convert the aligned photos and their dense depths to an LDI representation~\cite{shade1998layered}, in which layers are separated according to depth discontinuities, and apply RGBD inpainting in occluded regions as described below.

\begin{figure}
    \centering
    \includegraphics[width=\linewidth]{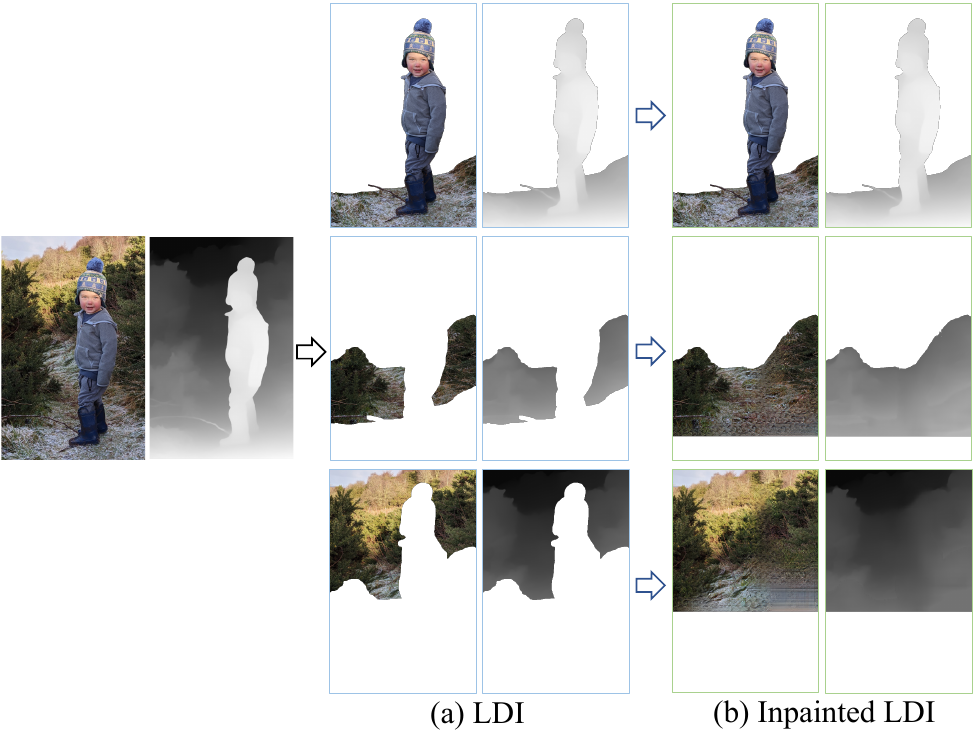}  \caption{\textbf{From an image to an inpainted LDI.} Given an input image and its estimated monocular depth~\cite{Ranftl2021}, we first apply agglomerative clustering~\cite{Maimon2005DataMA} to separate the RGBD image into multiple~(in this example 3) RGBDA layers as shown in (\textbf{a}), then perform context-aware color and depth inpainting~\cite{shih20203d} to obtain inpainted RGBDA layers~(\textbf{b}). 
    }
    \label{fig:layers}
\end{figure}

Prior methods for 3D photos iterate over all depth edges in an LDI to adaptively inpaint local regions using background pixels of the edge~\cite{shih20203d, Kopf2020OneS3}. However, we found this procedure to be computationally expensive and the output difficult to feed into a training pipeline.  More recently, Jampani et al.~\cite{jampani2021SLIDESI} employ a two-layer approach that would otherwise suit our requirements but is restricted in the number of layers. We therefore propose a simple, yet effective strategy for creating and inpainting LDIs that flow well into our learning-based pipeline. 
Specifically, we first perform agglomerative clustering~\cite{Maimon2005DataMA} in disparity space to separate the RGBD maps into different depth layers~(Fig.~\ref{fig:layers} (a)). 
We set a fixed \ignore{disparity} distance threshold above which clusters will not be merged, resulting in $2\sim5$ layers for an image.
We apply the clustering to the disparities of both images to obtain their LDIs, $\LDIa \triangleq \{ \ColorLa, \DepthLa \}^{L_0}_{l=1}$ and $\LDIb \triangleq \{ \ColorLb, \DepthLb \}^{L_1}_{l=1}$, \
where $\mathbf{C}^l$ and $\mathbf{D}^l$ represent the $l^\mathrm{th}$ color and depth layer respectively, and $L_0$ and $L_1$ denote the number of layers constructed from $\Ia$ and $\Ib$, respectively. Each color layer is an RGBA image, with the alpha channel indicating valid pixels in this layer.

Next, we apply depth-aware inpainting to each color and depth LDI layer in occluded regions. To inpaint missing contents in layer $l$, we treat all the pixels between the $l^\mathrm{th}$ layer and the farthest layer as the context region~(i.e., the region used as reference for inpainting), 
and exclude all irrelevant foreground pixels in layers nearer than layer $l$.
We set the rest of the $l^\mathrm{th}$ layer within a certain margin from existing pixels~(see supplement) to be inpainted. We keep only inpainted pixels whose depths are smaller than the maximum depth of layer $l$ so that inpainted regions do not mistakenly occlude layers farther than layer $l$. 
We adopt the pre-trained inpainting network from Shih~\textit{et al.}~\cite{shih20203d} to inpaint color and depth at each layer.
Fig.~\ref{fig:layers}~(b) shows an example of LDI layers after inpainting.
Note that we choose to inpaint the two LDIs up front rather than performing per-frame inpainting for each rendered novel view, as the latter would suffer from multi-view inconsistency due to the lack of a global representation for disoccluded regions.

\subsection{Space-time scene representation}

We now have inpainted color LDIs $\LDIa$ and $\LDIb$ for novel view synthesis. From each individual LDI, we could synthesize new views of the static scene. However, the LDIs alone do not model the scene motion between the two photos. To enable motion interpolation, we estimate 3D motion fields between the images. To do so, we first compute 2D optical flow between the two aligned images and perform a forward and backward consistency check to identify pixels with mutual correspondences. Given 2D mutual correspondences, we use their associated depth values to compute their 3D locations and lift the 2D optical flow to 3D scene flow, i.e., 3D translation vectors that displace each 3D point from one time to another. This process gives the scene flow for mutually visible pixels of the LDIs.

However, for pixels that do not have mutual correspondences, such as those occluded in the other view or those in the inpainted region, 3D correspondences are not well defined. To handle this issue, we leverage the fact that the scene flows are spatially smooth and propagate them from well-defined pixels to missing regions. In particular, for each pixel in $\LDIa$ with a corresponding point in $\LDIb$, we store its associated scene flow at its pixel location, resulting in scene flow layers initially containing only well-defined values for mutually visible pixels.
To inpaint the remaining scene flow, we perform a diffusion operation that iteratively applies a masked blur filter to each scene flow layer until all pixels in $\LDIa$ have scene flow vectors. 
We apply the same method to $\LDIb$ to obtain complete scene flow layers for the second LDI.
This process gives us complete forward and backward scene flows for every pixel in $\LDIa$ and $\LDIb$, respectively.

To render an image from a novel camera viewpoint and time with these two scene-flow-augmented LDIs,
one simple approach is to directly interpolate the LDI point locations to 
the target time according to their scene flow and splat RGB values to the target view. 
However, when using this method, we found that any small error in depth or scene flow can lead to noticeable artifacts. 
We therefore \ignore{use machine learning to} correct for such errors
by training a 2D feature extraction network that takes each inpainted LDI color layer $\mathbf{C}^l$ as input and produces a corresponding 2D feature map $\mathbf{F}^l$. These features encode local appearance of the scene and are trained to mitigate rendering artifacts introduced by inaccurate depth or scene flow and to improve overall rendering quality.
This step converts our inpainted color LDIs to feature LDIs $\FLDIa \triangleq \{ \FeatureLa, \DepthLa \}_{l=1}^{L_0}$,  $\FLDIb \triangleq \{\FeatureLb, \DepthLb\}_{l=1}^{L_1}$, both of which are augmented with scene flows. 
Finally, we lift all valid pixels for these feature LDIs into a pair of point clouds $\Ptsa \triangleq \{ (\Posa,  \Featurea, \sfa) \}$ and $\Ptsb \triangleq \{ (\Posb, \Featureb, \sfb) \}$, where each point is defined with 3D location $\mathbf{x}$, appearance feature $\mathbf{f}$, and 3D scene flow $\mathbf{u}$.

\subsection{Bidirectional splatting and rendering}

Given a pair of 3D feature point clouds $\Ptsa$ and $\Ptsb$, we wish to interpolate and render them to produce the image at a novel view and time $t$. 
Inspired by prior work~\cite{Niklaus_CVPR_2020_softsplat,Bao_CVPR_2019_dain}, we propose a depth-aware bidirectional splatting technique.
In particular, we first obtain the 3D location of every point (in both point clouds) at time $t$ by displacing it according to its associated scene flow scaled by $t$: $\Posat = \Posa + t \sfa$, $\Posbt = \Posb + (1 - t) \sfb$. The displaced points and their associated features from each direction~($0\rightarrow t$ or $1\rightarrow t$) are then separately splatted into the target viewpoint using differentiable point-based rendering~\cite{Wiles2020SynSinEV}, which results in a pair of rendered 2D feature maps $\FeatureMapat, \FeatureMapbt$ and depth maps $\Depthat, \Depthbt$. To combine the two feature maps and decode them to a final image, we linearly blend them based on spatial-temporal cues. Our general principles are: 1) if $t$ is closer to $0$ then $ \FeatureMapat$ should have a higher weight, and vice versa, and 2) for a 2D pixel, if its splatted depth $\Depthat$ from time $0$ is smaller then the depth $\Depthbt$ from time $1$, $\FeatureMapat$ should be favored more, and vice versa. Therefore, we compute a weight map to linearly blend the two feature and depth maps as follows:
\begin{align}
       \WeightMapt & = \frac{(1 - t) \cdot \exp(-\beta \cdot \Depthat)}{(1 - t) \cdot \exp(-\beta \cdot \Depthat) + t \cdot \exp(-\beta \cdot \Depthbt)} \\ 
    \FeatureMapt & = \WeightMapt \cdot \FeatureMapat + (\mathbf{1} - \WeightMapt) \cdot \FeatureMapbt \\
    \Deptht & = \WeightMapt \cdot \Depthat + (\mathbf{1} - \WeightMapt) \cdot \Depthbt.
\end{align} 
Here $\beta\in \mathbb{R}_+$ is a learnable parameter that controls contributions based on relative depth. 
Finally, $\FeatureMapt$ and $\Deptht$ are fed to a network that synthesizes the final color image.

\subsection{Training}

We train the feature extractor, image synthesis network, and the parameter $\beta$ on two video datasets to optimize the rendering quality, as described below.

\medskip
\noindent \textbf{Training datasets.}
To train our system, we ideally would use image triplets with known camera parameters, where each triplet depicts a dynamic scene from a moving camera, so that we can use two images as input and the third one (at an intermediate time and novel viewpoint) as ground truth. 
However, such data is difficult to collect at scale, since it either requires capturing dynamic scenes with synchronized multi-view camera systems, or running SfM on dynamic videos shot from moving cameras. The former requires a time-consuming setup and is difficult to scale to in-the-wild scenarios, while the latter cannot guarantee the accuracy of estimated camera parameters due to moving objects and potentially insufficient motion parallax. Therefore, we found that existing datasets of this kind are not sufficiently large or diverse for use as training data. Instead, we propose two sources of more accessible  
data for joint training of motion interpolation and view synthesis.

The first source contains video clips with small camera motions (unknown pose). We assume that the cameras are static and all pixel displacements are induced by scene motion. This type of data allows us to learn motion interpolation without the need for camera calibration.
The second source is video clips of static scenes with known camera motion. The camera motion of static scenes can be robustly estimated using SfM and such data gives us supervision for learning novel view synthesis. 
For the first source, we use Vimeo-90K~\cite{xue2019video}, a widely used dataset for learning frame interpolation. For the second source, we use the MannequinChallenge dataset~\cite{li2019learning}, which contains over 170K video frames of humans pretending to be statues captured from moving cameras, with corresponding camera poses estimated through SfM~\cite{zhou2018stereo}. Since the scenes in this dataset including people are (nearly) stationary, the estimated camera parameters are sufficiently accurate for our purposes. We mix these two datasets to train our model.

\smallskip
\noindent \textbf{Learnable components}. Our system consists of several modules: (a) monocular depth estimator, (b) color and depth inpainter, (c) 2D feature extractor, (d) optical flow estimator and (e) image synthesis network. We could conceptually train this whole system, but in practice we train only modules (c), (d), and (e), and use pretrained state-of-the-art models~\cite{Ranftl2021,shih20203d} for (a) and (b). This makes training 
less computationally expensive, and also avoids the need for the large-scale direct supervision required for learning high-quality depth estimation and RGBD inpainting networks.

\smallskip
\noindent \textbf{Training losses}. We train our system using image reconstruction losses. In particular, we minimize perceptual loss~\cite{lpips,johnson2016perceptual} and $l_1$ loss between the predicted and ground-truth images to supervise our networks.

\section{Experiments}

\subsection{Implementation details}
\label{sec:impl_detail}

For the feature extractor, we use ResNet34~\cite{he2016deep} truncated after \texttt{layer3} followed by two additional up-sampling layers to extract feature maps for each RGB layer, which we augment with a binary mask to indicate which pixels are covered (observed or inpainted) in that layer. For the image synthesis network, we adopt a 2D U-Net architecture. For the optical flow estimator we use a pre-trained RAFT network~\cite{teed2020raft} and fine-tune its weights during training.
We use Pytorch3D~\cite{ravi2020pytorch3d} for differentiable point cloud rendering. Rather than using a fixed radius for all points, we set the radius of a point proportionally to its disparity when rendering a target viewpoint. This prevents foreground objects from becoming semi-transparent due to gaps between samples when the camera zooms in. 

We train our system using Adam~\cite{Kingma2014AdamAM}, with base learning rates set to $10^{-4}$ for the feature extractor and image synthesis network, and $10^{-6}$ for the optical flow network~\cite{teed2020raft}.
We train our model on 8 NVIDIA V100 GPUs for 250k iterations for $\sim$ 3 days.
We decrease the learning rates exponentially during the optimization. Each training batch contains $8$ triplets randomly sampled from the Vimeo-90K~\cite{xue2019video} and MannequinChallenge datasets~\cite{li2019learning}. Within each triplet, the start and end images are used as input and the intermediate frame is used as ground truth. To train on MannequinChallenge, we must calibrate the monocular depth maps so that they align with the SfM point clouds. We estimate a disparity scale and shift for each depth map to minimize the MSE error between it and the depths of recovered SfM points. We discard sequences with large alignment errors during training. Please refer to the supplement for additional details.

\subsection{Baselines} %
\label{sec:quant_eval}

To our knowledge, there is no prior work that serves as a direct baseline for our new task of space-time view synthesis from the near-duplicate photos.
One might consider dynamic-NeRF approaches~\cite{Li2021NeuralSF, Xian2021SpacetimeNI, Pumarola2021DNeRFNR, Park2020DeformableNR} as baselines. However, these all require dense input views with known camera parameters and sufficient motion parallax, and thus do not apply to our scenario. Instead, as in NSFF~\cite{Li2021NeuralSF}, we found that we can combine individual methods to form baselines for our method. We describe three such baselines below. 

\smallskip
\noindent \textbf{Naive scene flow.}
As a simple baseline, we augment monocular depth with optical flow to get scene flow. 
Specifically, we first compute the monocular depths of the two views using DPT~\cite{Ranftl2021}, and lift them into 3D to get two colored point clouds. 
We then use 2D optical flows generated by RAFT~\cite{teed2020raft} to find pixels with mutual correspondences and compute their scene flows in the forward and backward directions. The two colored point clouds are then separately rendered to the target viewpoint at the intermediate time, producing two RGB images. Finally, we linearly blend the two rendered images based on the time $t$ to obtain the final view. Note that this baseline does not perform inpainting.  
 
\smallskip
\noindent \textbf{Frame interpolation $\rightarrow$ 3D photo}. 
Existing methods for frame interpolation and novel view synthesis can be combined to form a baseline for our task. Specifically, to synthesize an image at the novel time and viewpoint, we first adopt a state-of-the-art frame interpolation method,  XVFI~\cite{Sim2021XVFIEV}, to synthesize a frame at the intermediate time. We then apply 3D photo inpainting~\cite{shih20203d} to turn the interpolated frame into an inpainted LDI and render it from a desired viewpoint through a constructed mesh. For a fair comparison, we upgrade the 3D photo method to use the state-of-the-art monocular depth backbone DPT~\cite{Ranftl2021}, i.e., the same monocular depth predictor we use in our approach.
 
\smallskip
\noindent \textbf{3D photo $\rightarrow$ frame interpolation.} This baseline reverses the order of operations in the aforementioned method. First, we apply the 3D photo~\cite{shih20203d} to each of the near-duplicates and render them to the target viewpoint separately. We then apply XVFI~\cite{Sim2021XVFIEV} to these two rendered images to obtain a final view at intermediate time $t$.

\newcommand{\tablespace}{\,\,\,\,}
\newcommand{\halftablespace}{\,}
\setlength{\tabcolsep}{4pt}
\begin{table*}[t]
\centering
\small

\begin{tabular}{l |ccc | ccc}
\toprule
& \multicolumn{3}{c|}{NVIDIA Dynamic Scene~\cite{yoon2020novel}} & \multicolumn{3}{c}{UCSD Multi-View Video~\cite{Lin2021Deep3M}}
\\
Method &  PSNR$\uparrow$ & SSIM$\uparrow$ & LPIPS $\downarrow$ &  PSNR$\uparrow$ & SSIM$\uparrow$ & LPIPS$\downarrow$\\

\midrule
Naive Scene Flow & 19.34 & 0.681 & 0.177 
& 23.60 & 0.837 & 0.120 
\\

Frame Interpolation~\cite{Sim2021XVFIEV} $\rightarrow$ 3D Photo~\cite{shih20203d} & 21.01 & 0.676 & 0.189 
& 25.70 & 0.852 & 0.123 
\\

3D Photo~\cite{shih20203d} $\rightarrow$ Frame Interpolation~\cite{Sim2021XVFIEV} & 21.18 & 0.681  & 0.192 
& 25.96 & 0.858 & 0.126 
\\

Ours & $\mathbf{21.72}$ & $\mathbf{0.702}$ & $\mathbf{0.145}$ 
& $\mathbf{26.54}$ & $\mathbf{0.864}$ & $\mathbf{0.078}$  
\\
\bottomrule
\end{tabular}
\caption{\textbf{Quantitative comparisons of novel view and time synthesis.} Our method outperforms all the baselines in all error metrics. See Sec.~\ref{sec:quant_eval} for the descriptions of baselines.}
\label{table:results}  
\end{table*}
 
\subsection{Comparisons on public benchmarks}
\noindent \textbf{Evaluation datasets.}
We evaluate our method and baselines on two public multi-view dynamic scene datasets: the NVIDIA Dynamic Scenes Dataset~\cite{yoon2020novel} and the UCSD Multi-View Video Dataset~\cite{Lin2021Deep3M}. The NVIDIA dataset consists of 9 scenes involving more challenging human and non-human motions captured by 12 synchronized cameras at 60FPS. The UCSD dataset contains 96 multi-view videos of dynamic scenes, which capture diverse human interactions in outdoor environments. The videos are recorded by 10 synchronized action cameras at 120FPS. We run COLMAP~\cite{schonberger2016structure} on each of the multi-view videos~(masking out dynamic components using provided motion masks) to obtain camera parameters and sparse point clouds of the static scene contents.

\smallskip
\noindent \textbf{Experimental setup.}
To evaluate rendering quality, we sample a triplet~(two input and one target view) every $0.5$ seconds from the multi-view videos. In each triplet, we select the two input views to be at the same camera viewpoint and two frames apart, and the target view to be the middle frame at a nearby camera viewpoint. We compare the prediction with the ground truth at the same time and viewpoint. 
All methods we evaluate use monocular depths that are only predicted up to an unknown disparity scale and shift. To properly render images into the target viewpoint and compare with the ground truth, we need to obtain aligned depth maps that are consistent with the reconstructed scenes. Similar to Sec.~\ref{sec:impl_detail}, we align the predicted depths with the depth from SfM point clouds. Please refer to the supplement for more detail.

\smallskip
\noindent \textbf{Quantitative comparisons.}
We evaluate the rendering quality of each method using three standard error metrics: PSNR, SSIM and LPIPS~\cite{lpips}.
Tab.~\ref{table:results} shows comparisons between our method and the baselines. Our method consistently outperforms the baselines in all error metrics. In particular, our LPIPS scores are significantly better, suggesting better perceptual quality and photorealism of rendered images for our approach. Note that all the methods have relatively low PSNR/SSIM because these metrics are sensitive to pixel misalignment, and inaccurate geometry from monocular depth networks can cause the rendered images to not fully align with the ground truth. But since all methods use DPT~\cite{Ranftl2021} depths, this issue does not affect the relative comparisons.

\begin{figure}
    \centering
    \includegraphics[width=\linewidth]{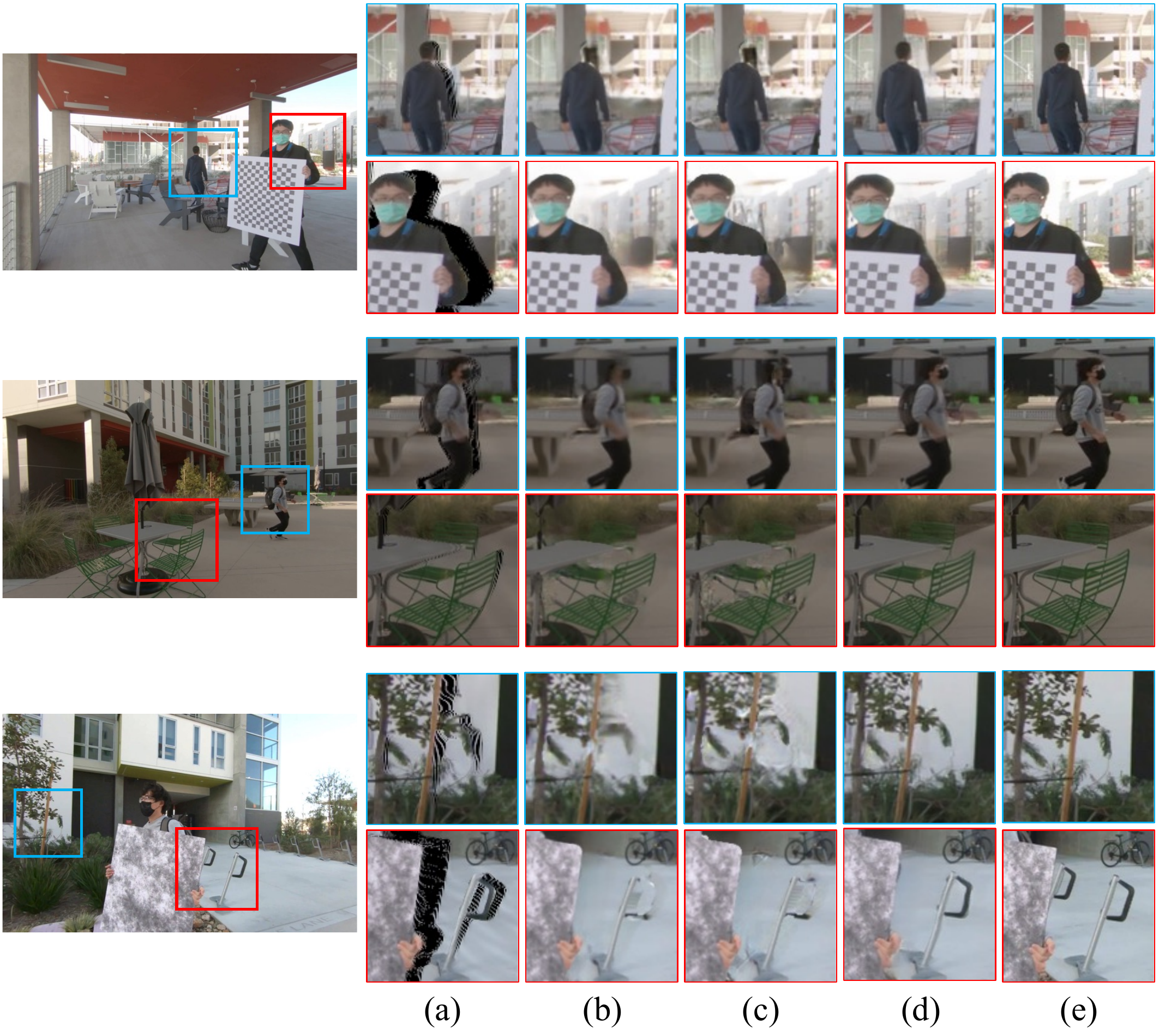}
    \caption{\textbf{Qualitative comparisons on the UCSD dataset~\cite{Lin2021Deep3M}}.
    From left to right are (\textbf{a}) naive scene flow,
    (\textbf{b}) frame interpolation~\cite{Sim2021XVFIEV} $\rightarrow$ 3D Photo~\cite{shih20203d}, (\textbf{c}) 3D Photo~\cite{shih20203d}$\rightarrow$ frame interpolation~\cite{Sim2021XVFIEV}, (\textbf{d}) our method, and (\textbf{e}) ground truth.
    }
    \label{fig:ucsd}
\end{figure}

\smallskip
\noindent \textbf{Qualitative comparisons.} We show qualitative comparisons on the UCSD dataset in Fig.~\ref{fig:ucsd}. Our method generates the fewest artifacts while preserving the most details in the scene.
The naive scene flow baseline produces noticeable holes. Applying 3D Photos and then frame interpolation leads to blurry disoccluded regions as the frame interpolator~\cite{Sim2021XVFIEV} is not trained to interpolate between inconsistently inpainted images. Applying frame interpolation and then 3D Photos leads to strong flickering artifacts due to inconsistent inpainting in each frame~(see supplement video).

\subsection{Comparisons on in-the-wild photos}
\begin{figure*}[h]
    \centering
    \includegraphics[width=\linewidth]{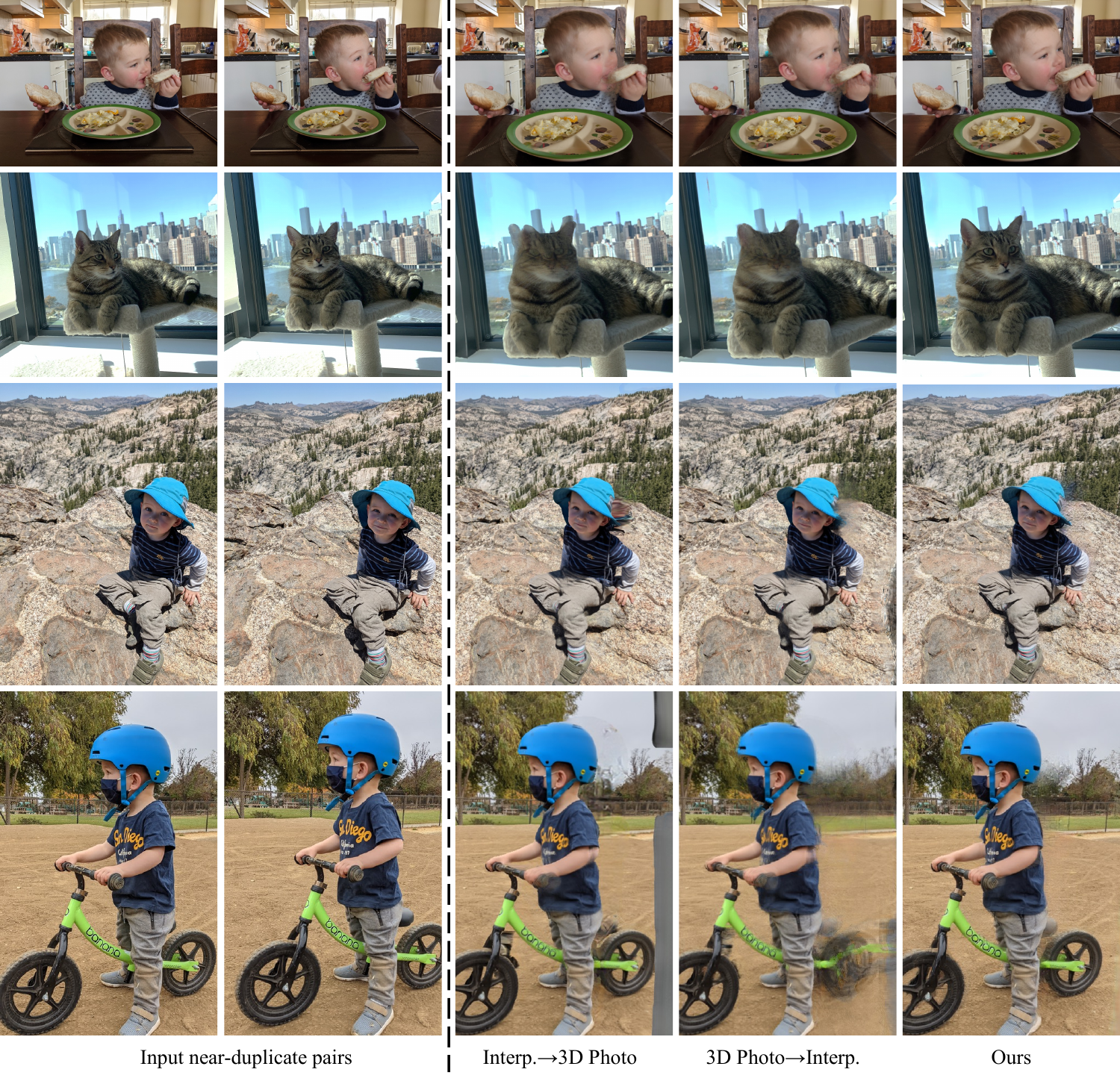}
    \caption{\textbf{Qualitative comparisons on in-the-wild photos.} Compared with the baselines, our approach produces more realistic views with significantly fewer visual artifacts, especially in moving or disoccluded regions. Please see the supplemental video for animated comparisons.
    }
    \label{fig:qualitative}
\end{figure*}

We also evaluate our approach and the baselines qualitatively on in-the-wild near-duplicate photos.
We collected these photos from our colleagues and their friends and families and obtained their consent to present these photos in this manuscript. We show comparisons of views generated by each method in Fig.~\ref{fig:qualitative}. In particular, we show two different kinds of camera motions, zooming in and tracking, and rendering a novel view at intermediate time $t=0.5$. Our method achieves overall better rendering quality with fewer visual artifacts, especially near moving objects and occlusion boundaries. We refer readers to the supplementary video for better visual comparisons of these generated 3D Moments.

\begin{table}[t]
    \centering
    \small
    \begin{tabular}{l ccc}
    \toprule
        & PSNR$\uparrow$ & SSIM$\uparrow$ & LPIPS$\downarrow$ \\
    \midrule
        No features & $21.16$ & $0.693$ & $0.173$ \\ 
        No inpainting & $21.33$ & $0.685$ & $\mathbf{0.145}$ \\
        No bidirectional & $21.56$ & $0.694$ & $0.151$ \\
        Full model Ours & $\mathbf{21.72}$ & $\mathbf{0.702}$ & $\mathbf{0.145}$ \\ 
    \bottomrule
    \end{tabular}
    \caption{\textbf{Ablation studies on the NVIDIA dataset~\cite{yoon2020novel}}. Each component of our system leads to an increase in rendering quality. 
    }
    \label{tab:ablation}
\end{table}

\subsection{Ablations and analysis}
\noindent \textbf{Ablation studies.}
We conduct ablation studies to justify our design choices, as shown in Tab.~\ref{tab:ablation}. For ``No features'', instead of learning features we directly use RGB colors from the input photos to splat and render novel views.
For ``No inpainting'', we train the system without inpainting color and depth in our LDIs and rely on the image synthesis network to fill in disoccluded regions in each rendered view separately~(prone to temporal inconsistency). For ``No bidirectional warping'', we use only single-directional scene flow from time $0$ to time $1$.

\smallskip
\noindent \textbf{Performance.} Our method can be applied to new near-duplicate photo pairs without requiring test-time optimization. We test our runtime on an NVIDIA V100 GPU. Given a duplicate pair of images with resolution 768 $\times$ 576, it takes $4.48$s to build LDIs, extract feature maps, and build the 3D feature scene flow. These operations are performed once for each duplicate pair. The projection-and-image-synthesis stage takes $0.71$s to render each output frame.

\section{Discussion and Conclusion}

We presented a new task of creating 3D Moments from near-duplicate photos, allowing simultaneous view extrapolation and motion interpolation for a dynamic scene. We propose a new system for this task that models the scene as a pair of feature LDIs augmented with scene flows. By training on both posed and unposed video datasets, our method is able to produce photorealistic space-time videos from the near-duplicate pairs without substantial visual artifacts or temporal inconsistency. Experiments show that our approach outperforms the baseline methods both quantitatively and qualitatively on the tasks of space-time view synthesis.

\noindent \textbf{Limitations and future work.}
Our method inherits some limitations of monocular depth and optical flow methods. Our method does not work well for photos with complex scene geometry or semi-transparent objects. In addition, our method tends to fail in the presence of large and non-linear motions as well as challenging self-occlusions, such as hands. Please refer to the supplementary video for failure cases. Future work includes designing an automatic selection scheme for photo pairs suitable for 3D Moment creation, automatically detecting failures, better modeling of large or non-linear motions, and extending the current method to handle more than two near-duplicate photos.

\noindent 
\textbf{Acknowledgements.} We thank Richard Tucker, Tianfan Xue, Andrew Liu, Jamie Aspinall, Fitsum Reda and Forrester Cole for help, discussion and support.

{\small
\bibliographystyle{ieee_fullname}
\bibliography{refs}
}

\end{document}